\documentclass{article}

\PassOptionsToPackage{numbers}{natbib}

\usepackage[final]{neurips_2021}

\usepackage[utf8]{inputenc} %
\usepackage[T1]{fontenc}    %
\usepackage{hyperref}       %
\usepackage{url}            %
\usepackage{booktabs}       %
\usepackage{amsfonts}       %
\usepackage{nicefrac}       %
\usepackage{microtype}      %
\usepackage{xcolor}         %
\usepackage{graphicx}
\usepackage{amsmath}
\usepackage{commath}
\usepackage{subcaption}
\usepackage{caption}
\usepackage{wrapfig}
\usepackage{algorithm}
\usepackage{algorithmic}

\title{Arbitrary Conditional Distributions with Energy}

\author{%
  Ryan~R.~Strauss \\
  Department of Computer Science\\
  UNC at Chapel Hill\\
  Chapel Hill, NC 27514 \\
  \texttt{rrs@cs.unc.edu} \\
  \And
  Junier~B.~Oliva \\
  Department of Computer Science\\
  UNC at Chapel Hill\\
  Chapel Hill, NC 27514 \\
  \texttt{joliva@cs.unc.edu} \\
}

\begin{document}

\maketitle

\begin{abstract}
Modeling distributions of covariates, or \textit{density estimation}, is a core challenge in unsupervised learning. However, the majority of work only considers the \textit{joint} distribution, which has limited utility in practical situations. A more general and useful problem is \textit{arbitrary conditional density estimation}, which aims to model \textit{any} possible conditional distribution over a set of covariates, reflecting the more realistic setting of inference based on prior knowledge. We propose a novel method, Arbitrary Conditioning with Energy (ACE), that can simultaneously estimate the distribution $p(\mathbf{x}_u \mid \mathbf{x}_o)$ for all possible subsets of unobserved features $\mathbf{x}_u$ and observed features $\mathbf{x}_o$. ACE is designed to avoid unnecessary bias and complexity --- we specify densities with a highly expressive energy function and reduce the problem to only learning one-dimensional conditionals (from which more complex distributions can be recovered during inference). This results in an approach that is both simpler and higher-performing than prior methods. We show that ACE achieves state-of-the-art for arbitrary conditional likelihood estimation and data imputation on standard benchmarks.
\end{abstract}

\section{Introduction}

Density estimation, a core challenge in machine learning, attempts to learn the probability density of some random variables given samples from their true distribution. The vast majority of work on density estimation focuses on the \textit{joint} distribution $p(\mathbf{x})$ \cite{goodfellow2014generative,dinh2016density,papamakarios2017masked,grathwohl2018ffjord,oliva2018transformation,nash2019autoregressive,fakoor2020trade}, i.e.,~the distribution of all variables taken together. While the joint distribution can be useful (e.g.,~to learn the distribution of pixel configurations that represent human faces), it is limited in the types of predictions it can make. We are often more interested in \textit{conditional} probabilities, which communicate the likelihood of an event \textit{given} some prior information. For example, given a patient's medical history and symptoms, a doctor determines the likelihoods of different illnesses and other patient attributes. Conditional distributions are often more practical since real-world decisions are nearly always informed by prior information. 

However, we often do not know ahead of time \textit{which} conditional distribution (or distributions) will be of interest. That is, we may not know which features will be known (observed) or which will be inferred (unobserved). For example, not every patient that a doctor sees will have had the same tests performed (i.e.,~have the same observed features). A na\"ive approach requires building an exponential number of models to cover all possible cases (one for every conditional distribution), which quickly becomes intractable. Thus, an intelligent system needs to understand the intricate conditional dependencies between all arbitrary subsets of covariates, and it must do so with a single model to be practical.

In this work, we consider the problem of learning the conditional distribution $p(\mathbf{x}_u \mid \mathbf{x}_o)$ for any arbitrary subsets of unobserved variables $\mathbf{x}_u \in \mathbb{R}^{|u|}$ and observed variables $\mathbf{x}_o \in \mathbb{R}^{|o|}$, where $u,o \subseteq \{1, \dots , d\}$ and $o \cap u = \emptyset$.
We propose a method, Arbitrary Conditioning with Energy (ACE), that can assess any conditional distribution over any subset of random variables, using a single model. ACE is developed with an eye for simplicity --- we reduce the arbitrary conditioning problem to the estimation of one-dimensional conditional densities (with arbitrary observations), and we represent densities with an energy function, which fully specifies unnormalized distributions and has the freedom to be instantiated as any arbitrary neural network.

We evaluate ACE on benchmark datasets and show that it outperforms current methods for arbitrary conditional/marginal density estimation. ACE remains effective when trained on data with missing values, making it applicable to real-world datasets that are often incomplete, and we find that ACE scales well to high-dimensional data. Also, unlike some prior methods (e.g.,~\cite{li2020acflow}), ACE can naturally model data with both continuous and discrete values.

Our contributions are as follows: 1) We develop the first energy-based approach to arbitrary conditional density estimation, which eliminates restrictive biases (e.g.~normalizing flows, Gaussian mixtures) imposed by common alternatives. 2) We empirically demonstrate that ACE is state-of-the-art for arbitrary conditional density estimation and data imputation. 3) We find that complicated prior approaches can be easily outperformed with a simple scheme that uses mixtures of Gaussians and fully-connected networks.

\section{Previous Work}

Several methods have been previously proposed for arbitrary conditioning. Sum-Product Networks are specially designed to only contain sum and product operations and can produce arbitrary conditional or marginal likelihoods \cite{poon2011sum,butz2019deep}. The Universal Marginalizer trains a neural network with a cross-entropy loss to approximate the marginal posterior distributions of all unobserved features conditioned on the observed ones \cite{douglas2017universal}. VAEAC is an approach that extends a conditional variational autoencoder by only considering the latent codes of unobserved dimensions \cite{ivanov2018variational}, and NeuralConditioner uses adversarial training to learn each conditional distribution \cite{belghazi2019learning}. DMFA uses factor analysis to have a neural network output the parameters of a conditional Gaussian density for the missing features given the observed ones \cite{przewikezlikowski2020estimating}. The current state-of-the-art is ACFlow, which extends normalizing flow models to handle any subset of observed features \cite{li2020acflow}.

Unlike VAEAC, ACE does not suffer from mode collapse or blurry samples. ACE is also able to provide likelihood estimates, unlike NeuralConditioner which only produces samples. DMFA is limited to learning multivariate Gaussians, which impose bias and are harder to model than one-dimensional conditionals. While ACFlow can analytically produce normalized likelihoods and samples, it is restricted by a requirement that its network consist of bijective transformations with tractable Jacobian determinants. Similarly, Sum-Product Networks have limited expressivity due to their constraints. ACE, on the other hand, exemplifies the appeal of energy-based methods as it has no constraints on the parameterization of the energy function.

Energy-based methods have a wide range of applications within machine learning \cite{lecun2006tutorial}, and recent work has studied their utility for density estimation. Deep energy estimator networks \cite{saremi2018deep} and Autoregressive Energy Machines \cite{nash2019autoregressive} are both energy-based models that perform density estimation. However, both of these methods are only able to estimate the joint distribution.

Much of the previous work on density estimation relies on an autoregressive decomposition of the joint density according to the chain rule of probability. Often, the model only considers a single (arbitrary) ordering of the features \cite{nash2019autoregressive,oord2016conditional}. \citet{uria2014deep} proposed a method for assessing joint likelihoods based on any arbitrary ordering, where they use masked network inputs to effectively share weights between a combinatorial number of models.
\citet{germain2015made} also consider a shared network for joint density estimation with a constrained architecture that enforces the autoregressive constraint in joint likelihoods. 
In this work, we construct an order-agnostic weight-sharing technique not for joint likelihoods, but for arbitrary conditioning.
Moreover, we make use of our weight-sharing scheme to estimate likelihoods with an energy based approach, which avoids the limitations of the parametric families used previously (e.g.,~mixtures of Gaussians  \cite{uria2014deep}, or Bernoullis \cite{germain2015made}).
Query Training \cite{lazaro2020query} is a method for answering probabilistic queries. It also takes the approach of computing one-dimensional conditional likelihoods but does not directly pursue an autoregressive extension of that ability.

The problem of imputing missing data has been well studied, and there are several approaches based on classic machine learning techniques such as $k$-nearest neighbors \cite{troyanskaya2001missing}, random forests \cite{stekhoven2012missforest}, and autoencoders \cite{gondara2018mida}. More recent work has turned to deep generative models. GAIN is a generative adversarial network (GAN) that produces imputations with the generator and uses the discriminator to discern the imputed features \cite{yoon18gain}. Another GAN-based approach is MisGAN, which learns two generators to model the data and masks separately \cite{li2019misgan}. MIWAE adapts variational autoencoders by modifying the lower bound for missing data and produces imputations with importance sampling \cite{mattei2019miwae}. ACFlow can also perform imputation and is state-of-the-art for imputing data that are missing completely at random (MCAR) \cite{li2020acflow}.

While it is not always the case that data are missing at random, the opposite case (i.e.,~missingness that depends on unobserved features' values) can be much more challenging to deal with \cite{fielding2008simple}. Like many data imputation methods, we focus on the scenario where data are MCAR, that is, where the likelihood of being missing is independent of the covariates' values.

\section{Background}

\subsection{Arbitrary Conditional Density Estimation}

A probability density function (PDF), $p(\mathbf{x})$, outputs a nonnegative scalar for a given vector input $\mathbf{x} \in \mathbb{R}^d$ and satisfies $\int p(\mathbf{x}) \dif \mathbf{x} = 1$. 
Given a dataset $\mathcal{D} = \{ \mathbf{x}^{(i)} \}_{i=1}^N$ of \textit{i.i.d.}~samples drawn from an unknown distribution $p^*(\mathbf{x})$, the object of density estimation is to find a model that best approximates the function $p^*$. Modern approaches generally rely on neural networks to 
directly parameterize the approximated PDF.

Arbitrary conditional density estimation is a more general task where we estimate the conditional density $p^*(\mathbf{x}_u \mid \mathbf{x}_o)$ for all possible subsets of observed features \mbox{$o \subset \{1, \dots, d\}$}  (i.e.,~features whose values are known) and corresponding subset of unobserved features \mbox{$u \subset \{1, \dots , d\}$} such that $o$ and $u$ do not intersect. Here $\mathbf{x}_o \in \mathbb{R}^{|o|}$ and $\mathbf{x}_u \in \mathbb{R}^{|u|}$. The estimation of joint or marginal likelihoods is recovered when $o$ is the empty set. Note that marginalization to obtain $p(\mathbf{x}_o) = \int p(\mathbf{x}_u, \mathbf{x}_o) \dif \mathbf{x}_u$ (and hence $p(\mathbf{x}_u \mid \mathbf{x}_o) = \frac{p(\mathbf{x}_u, \mathbf{x}_o)}{p(\mathbf{x}_o)}$) is intractable in performant generative models like normalizing flow based models \cite{dinh2016density,li2020acflow}; thus, we propose a weight-sharing scheme below.

\subsection{Energy-Based Models}

Energy-based models capture dependencies between variables by assigning a nonnegative scalar \textit{energy} to a given arrangement of those variables, where energies closer to zero indicate more desirable configurations \cite{lecun2006tutorial}. Learning consists of finding an energy function that outputs low energies for correct values. We can frame density estimation as an energy-based problem by writing likelihoods as a Boltzmann distribution
\begin{equation} \label{eq:energy-dist}
    p(\mathbf{x}) = \frac{e^{- \mathcal{E}(\mathbf{x})}}{Z}, \qquad Z = \int e^{- \mathcal{E}(\mathbf{x})} \dif \mathbf{x}
\end{equation}
where $\mathcal{E}$ is the energy function, $e^{- \mathcal{E}(\mathbf{x})}$ is the unnormalized likelihood, and $Z$ is the normalizer.

Energy-based models are appealing due to their relative simplicity and high flexibility in the choice of representation for the energy function. This is in contrast to other common approaches to density estimation such as normalizing flows \cite{dinh2016density,li2020acflow}, which require invertible transformations with Jacobian determinants that can be computed efficiently. Energy functions are also naturally capable of representing non-smooth distributions with low-density regions or discontinuities.

\section{Arbitrary Conditioning with Energy}

We are interested in approximating the probability density $p(\mathbf{x}_u \mid \mathbf{x}_o)$ for any arbitrary sets of unobserved features $\mathbf{x}_u$ and observed features $\mathbf{x}_o$. We approach this by decomposing likelihoods into products of one-dimensional conditionals, which makes the learned distributions much simpler. This concept is a basic application of the chain rule of probability, but has yet to be thoroughly exploited for arbitrary conditioning. During training, ACE estimates distributions of the form $p(x_{u^\prime_i} \mid \mathbf{x}_{o^\prime})$, where $x_{u^\prime_i}$ is a scalar. During inference, more complex distributions can then be recovered with an autoregressive decomposition: \mbox{$p(\mathbf{x}_u \mid \mathbf{x}_o) = \prod_{i = 1}^{|u|} p(x_{u^\prime_i} \mid \mathbf{x}_{o\, \cup\, u^\prime_{<i}})$}, where $u^\prime$ is an arbitrary permutation of $u$ and $ u^\prime_{<i} = \{u^\prime_1, \dots, u^\prime_{i-1}\}$. This approach is appealing because the estimated distributions are over a one-dimensional domain, allowing one to use a myriad of flexible estimators.

We adopt an energy-based approach (similar to AEMs \cite{nash2019autoregressive}), which affords a large degree of flexibility in modeling the exponentially many conditional distributions at hand --- we are free to represent the energy function with an arbitrary, and highly expressive, neural network that directly outputs unnormalized likelihoods. This contrasts with the current state-of-the-art for arbitrary conditional density estimation, which is limited by normalizing flow transformations \cite{li2020acflow}. Energy functions are highly expressive and naturally model complex (i.e.,~multimodal, non-smooth) densities, as they avoid a parametric-family on the shape of the distribution. Our main contribution is a method, Arbitrary Conditioning with Energy (ACE), for computing arbitrary conditional likelihoods with energies, one dimension at a time.\footnote{An implementation of ACE is available at \url{https://github.com/lupalab/ace}.}

\subsection{Decomposing Densities}

\begin{figure}[t]
    \centering
    \begin{subfigure}[b]{0.50\linewidth}
        \centering
        \includegraphics[width=\linewidth]{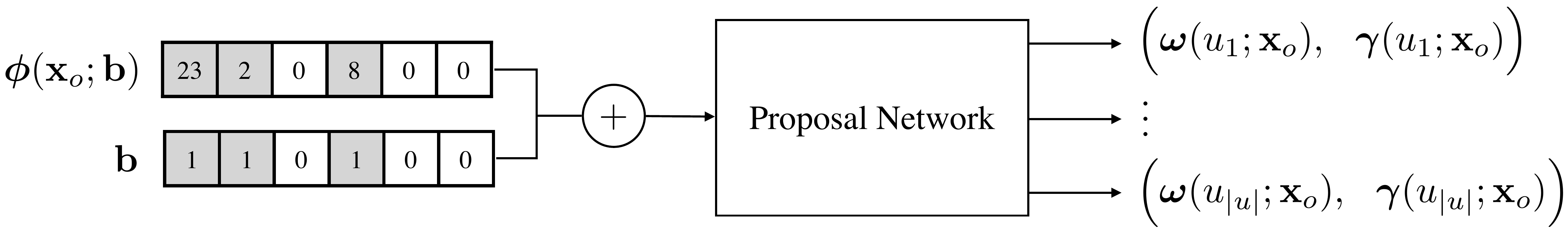}
        \caption{Proposal Network}
        \label{fig:proposal-network}
    \end{subfigure}
    \hfill
    \begin{subfigure}[b]{0.46\linewidth}
        \centering
        \includegraphics[width=\linewidth]{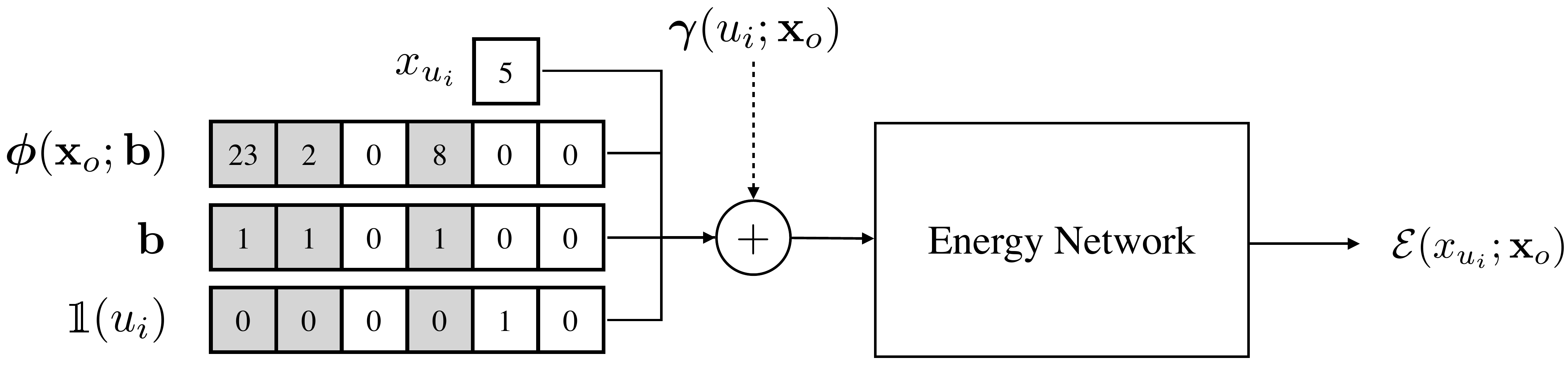}
        \caption{Energy Network}
        \label{fig:energy-network}
    \end{subfigure}
    \caption{Overview of the networks used in ACE. The plus symbol refers to concatenation.}
    \label{fig:networks}
\end{figure}

We decompose the arbitrary conditioning task into $1d$-domain arbitrary conditional estimation problems. This follows from the chain rule of probability, which allows us to write
\begin{equation} \label{eq:chain-rule}
    p(\mathbf{x}_u \mid \mathbf{x}_o) = \prod_{i = 1}^{|u|} p\left(x_{u^\prime_i} \mid \mathbf{x}_{o\, \cup\, u^\prime_{<i}}\right),
\end{equation}
where $u^\prime$ is an arbitrary permutation of $u$, $x_{u^\prime_i}$ is the $i^{\text{th}}$ unobserved feature given by $u^\prime$, and $u^\prime_{<i} = \{u^\prime_1, \dots, u^\prime_{i-1}\}$. The one-dimensional conditionals in \autoref{eq:chain-rule} are themselves arbitrary conditionals due to the choice of permutation. Thus, the arbitrary conditioning problem may be reduced to one-dimensional arbitrary conditional estimation. If we can compute any likelihood of the form $p(x_{i} \mid \mathbf{x}_{o^\prime})$ for $o^\prime \subset \{1, \ldots, d\}$, we can compute any distribution over the features. This includes all possible conditional distributions, marginal distributions, and the joint distribution. Although the reduction of arbitrary conditioning to 1$d$ conditional estimation is seemingly straightforward, this is, to the best of our knowledge, the first work to leverage this formulation.

We train our model non-autoregressively, to output the likelihood $p(x_{i} \mid \mathbf{x}_{o^\prime})$ for any $x_{i}$ and $\mathbf{x}_{o^\prime}$, where $i\in \{1, \ldots, d\} \setminus o^\prime$. During inference, an autoregressive procedure that repeatedly applies the chain rule can be used to compute likelihoods of the form $p(\mathbf{x}_u \mid \mathbf{x}_o)$. In practice, this consists of simply adding the previously considered unobserved dimension to the observed set before moving to the next step in the chain.\footnote{ One could also implement this with a masked network architecture (e.g., MADE \cite{germain2015made}) that can more efficiently compute autoregressive likelihoods. We choose not to do this for simplicity and to enable the use of arbitrary, and more expressive, architectures.}

\subsection{Likelihoods from Energies}

We express likelihoods in terms of energies by modifying \autoref{eq:energy-dist} to write
\begin{equation} \label{eq:likelihood-energy}
    p(x_{u_i} \mid \mathbf{x}_o) = \frac{e^{- \mathcal{E}(x_{u_i}; \mathbf{x}_o)}}{Z_{u_i; \mathbf{x}_o}}~,
\end{equation}
and we choose to represent the energy function as a neural network. In order to avoid learning a different network for each conditional, we adopt a weight-sharing scheme (e.g.~\cite{uria2014deep,ivanov2018variational,li2020acflow}). That is, we use a bitmask \mbox{$\mathbf{b} \in \{0, 1\}^d$} that indicates which features are observed and define a zero-imputing function $\boldsymbol{\phi}(\mathbf{x}_o;\mathbf{b})$ that returns a $d$-dimensional vector where unobserved features are replaced with zeros (see \autoref{fig:zero-imputation} in Appendix). These are then used as inputs to the energy network (see \autoref{fig:energy-network}).

\paragraph{Approximating normalized likelihoods.}
We can use \autoref{eq:likelihood-energy} to compute normalized likelihoods, but only if the normalizing constant $Z_{u_i; \mathbf{x}_o}$ is known. Directly computing the normalizer is intractable in general. However, importance sampling can be used to obtain an estimate \cite{bishop2006pattern}. \citet{nash2019autoregressive} show that this technique is sufficiently accurate when considering one-dimensional conditionals  for joint density estimation, and we adapt the approach for arbitrary conditioning. Assuming access to a proposal distribution $q(x_{u_i} \mid \mathbf{x}_o)$ which is similar to the target distribution, we approximate $Z_{u_i; \mathbf{x}_o}$ as
\begin{align}
    Z_{u_i; \mathbf{x}_o} &= \int e^{- \mathcal{E}(x_{u_i}; \mathbf{x}_o)} \dif x_{u_i} = \int \dfrac{e^{- \mathcal{E}(x_{u_i}; \mathbf{x}_o)}}{q(x_{u_i} \mid \mathbf{x}_o)} q(x_{u_i} \mid \mathbf{x}_o) \dif x_{u_i} \\
    &\approx \dfrac{1}{S} \sum_{s=1}^S \dfrac{e^{- \mathcal{E}(x_{u_i}^{(s)}; \mathbf{x}_o)}}{q(x_{u_i}^{(s)} \mid \mathbf{x}_o)}, \quad x_{u_i}^{(s)} \sim q(x_{u_i} \mid \mathbf{x}_o)~. \label{eq:importance-weight}
\end{align}

For some problems, we may have access to a good proposal distribution ahead of time, but in general, we can learn one in parallel with the energy network. We note that importance sampling is also more computationally efficient than the simpler alternative of using points from a grid. Since distributions may be non-smooth and span a large domain, fewer samples from a good proposal distribution will be able to obtain as good of an estimate as a grid that has many more points, which in turn will require fewer evaluations of the energy network. Furthermore, generating samples from the proposal is very efficient, as it only requires one neural network evaluation to obtain the parameters of a tractable parametric distribution, which can then be easily sampled.

\subsection{Training ACE}
\label{sec:training}

We learn the proposal distribution alongside the energy function by having a neural network output the parameters of a tractable parametric distribution. For the proposal network, we again share weights between all conditionals as is done with the energy network. The proposal network accepts a concatenation of $\mathbf{b}$ and  $\boldsymbol{\phi}(\mathbf{x}_o;\mathbf{b})$ as input, and it outputs the parameters $\boldsymbol{\omega}(u_i;\mathbf{x}_o)$ for a mixture of Gaussians for each unobserved dimension $u_i$. The proposal network also outputs a latent vector, $\boldsymbol{\gamma}(u_i;\mathbf{x}_o)$, for each unobserved dimension, which is used as input to the energy network in order to enable weight sharing between the proposal and energy networks \cite{nash2019autoregressive}. Using a different latent vector for each feature allows the latent vectors to represent the fact that different unobserved features may depend on $\mathbf{x}_o$ in different ways.

We then estimate the normalizing constants in \autoref{eq:likelihood-energy} using importance sampling:
\begin{equation} \label{eq:normalizer-estimate}
    \hat{Z}_{u_i; \mathbf{x}_o} = \dfrac{1}{S} \sum_{s=1}^S \dfrac{e^{- \mathcal{E}(x_{u_i}^{(s)}; \mathbf{x}_o;\boldsymbol{\gamma}(u_i;\mathbf{x}_o))}}{q(x_{u_i}^{(s)} \mid \boldsymbol{\omega}(u_i;\mathbf{x}_o))}
\end{equation}
where $x_{u_i}^{(s)}$ is sampled from $q(x_{u_i} \mid \boldsymbol{\omega}(u_i;\mathbf{x}_o))$. This in turn leads to the following approximation of the log-likelihood of $x_{u_i}$ given $\mathbf{x}_o$:
\begin{equation} \label{eq:model-estimate}
    \log p(x_{u_i} \mid \mathbf{x}_o) \approx -\mathcal{E}(x_{u_i}; \mathbf{x}_o;\boldsymbol{\gamma}(u_i;\mathbf{x}_o)) - \log \hat{Z}_{u_i; \mathbf{x}_o} ,
\end{equation}
where we use abbreviated notation in the previous two equations and omit the bitmask $\mathbf{b}$ for greater readability. Refer to \autoref{fig:networks} for the precise inputs of each network.\footnote{It is not strictly necessary for the energy network to include $\mathbf{x}_o$ as input, since the latent vectors can encode that information. However, doing so acts like a sort of skip-connection, which often helps with deeper networks.}

Since \autoref{eq:chain-rule} gives us a way to autoregressively compute $p(\mathbf{x}_u \mid \mathbf{x}_o)$ as a chain of one-dimensional conditionals, we only concern ourselves with learning \mbox{$p(x_{u_i} \mid \mathbf{x}_o)$} for arbitrary $u_i$ and $\mathbf{x}_o$. Thus, for a given data point $\mathbf{x}$, we randomly partition it into $\mathbf{x}_o$ and $\mathbf{x}_u$ and jointly optimize the proposal and energy networks with the maximum-likelihood objective
\begin{equation} \label{eq:objective}
    J(\mathbf{x}_o; \mathbf{x}_u; \boldsymbol{\theta}) = \sum_{i=1}^{|u|} \log p(x_{u_i} \mid \mathbf{x}_o) + \sum_{i=1}^{|u|} \log q(x_{u_i} \mid \mathbf{x}_o),
\end{equation}
where $\boldsymbol{\theta}$ holds the parameters of both the energy and proposal networks. Because we want to optimize the proposal and energy distributions independently,
gradients are stopped on proposal samples and proposal likelihood before they are used in \autoref{eq:normalizer-estimate}. We note that optimizing \autoref{eq:objective} can be interpreted as a stochastic approximation to optimizing the full autoregressive likelihoods $p(\mathbf{x}_u \mid \mathbf{x}_o)$ (i.e., \autoref{eq:chain-rule}). For a given data point, we are choosing to optimize individual $1d$-conditionals from numerous hypothetical autoregressive chains, rather than all of the $1d$-conditionals from a single autoregressive chain. That is, each random set of observed values that is encountered during training (in \autoref{eq:objective}) can be seen as a single factor in the product in \autoref{eq:chain-rule} for some $p(\mathbf{x}_{u} \mid \mathbf{x}_o)$. Over the course of training though, the same conditionals are ultimately being evaluated in either case.

The negative of \autoref{eq:objective} is minimized with Adam \cite{kingma2014adam} over a set of training data, where observed and unobserved sets are selected at random for each minibatch (see \autoref{sec:experiments}). In some cases, we found it useful to include a regularization term in the loss that penalizes the energy distribution for large deviations from the proposal distribution. We use the mean-square error (MSE) between the proposal likelihoods and energy likelihoods as a penalty, with gradients stopped on the proposal likelihoods in the error calculation. The coefficient of this term in the loss is a hyperparameter.

\subsection{Inference}

\subsubsection{Likelihoods}
\label{sec:likelihoods}

Recall that our model learns one-dimensional conditionals: $p(x_{u_i} \mid \mathbf{x}_o)$.
Thus, to obtain a complete likelihood for $\mathbf{x}_u$, we employ an autoregressive application of the chain rule (see \autoref{eq:chain-rule}). The pseudocode for this procedure is presented in the Appendix (see \autoref{alg:likelihood}). Since the values of $\mathbf{x}_u$ are known ahead of time, each one-dimensional conditional can be evaluated in parallel as a batch, allowing the likelihood $p(\mathbf{x}_u \mid \mathbf{x}_o)$ to be computed efficiently.

Importantly, the order in which each unobserved dimension is evaluated does not matter. As argued in previous work \cite{uria2014deep,germain2015made}, this can be considered advantageous, since we can effectively leverage multiple orderings at test time to obtain an ensemble of models. However, this does incur extra computational cost during inference. We note that a model which perfectly captures the true densities would give consistent likelihoods for all possible orderings (thus evaluating only one ordering would suffice). However, current order-agnostic methods (such as ACE or \cite{uria2014deep,germain2015made}) do not inherently satisfy this constraint. In the Appendix, we study how these inconsistencies can be mitigated in ACE, but we leave a true solution to this challenge for future work.

\subsubsection{Imputing}
\label{sec:means}

Sampling allows us to obtain multiple possible values for the unobserved features that are diverse and realistic.\footnote{We describe how to generate samples from ACE in the Appendix.} However, these are not always the primary goals. For example, in the case of data imputation, we may only want a single imputation that aims to minimize some measure of error (see \autoref{sec:imputation}). Thus, rather than imputing true samples, we might prefer to impute the mean of the learned distribution. In this case, we forego autoregression and directly obtain the mean of each distribution $p(x_{u_i} \mid \mathbf{x}_o)$ with a single forward pass. Analytically computing the mean of the proposal distribution is straightforward since we are working with a mixture of Gaussians. We estimate the mean of the energy distribution via importance sampling:
\begin{equation}
    \mathbb{E} \left[ x_{u_i} \right] \approx \sum_{s=1}^S \dfrac{r_s}{\sum_j r_j} x_{u_i}^{(s)}, \qquad r_s = \dfrac{p(x_{u_i} \mid \mathbf{x}_o)}{q(x_{u_i} \mid \mathbf{x}_o)}
\end{equation}
where $x_{u_i}^{(s)}$ is sampled from $q(x_{u_i} \mid \mathbf{x}_o)$. It is worth noting that imputing the mean ignores dependencies between features in the unobserved set, so for some applications, other methods of imputing (such as multiple imputation by drawing samples) may make more sense.

\subsubsection{Heterogenous Data}

Prior approaches to arbitrary conditioning have to make restrictive assumptions when modeling arbitrary dependencies between continuous and discrete covariates. VAEAC \cite{ivanov2018variational}, for example, makes an assumption of conditional independence given a latent code. ACFlow \cite{li2020acflow} is \emph{not} directly applicable to discrete data given its use of the change of variable theorem. On the other hand, ACE can naturally model arbitrary dependencies between continuous and discrete covariates without any assumptions. In this setting, the proposal network outputs categorical logits for discrete features, as opposed to Gaussian mixture parameters. These logits can themselves be interpreted as energies, and we don't need to learn an additional energy function for the discrete features. Similarly, ACE could be applied to multimodal data in a straightforward fashion by conditioning the proposal distributions and energy function on a fused multimodal latent representation (which could be freely learned by arbitrary neural networks).

\section{Experiments}
\label{sec:experiments}

\subsection{Real-valued UCI Data}

We first evaluate ACE on real-valued tabular data. Specifically, we consider the benchmark UCI repository datasets described by \citet{papamakarios2017masked} (see \autoref{tab:datasets} in the Appendix).

Unlike other approaches to density estimation that require particular network architectures \cite{germain2015made,dinh2016density,nash2019autoregressive,li2020acflow}, ACE has no such restrictions. Thus, we use a simple fully-connected network with residual connections \cite{he2016deep,he2016identity} for both the energy network and proposal network. This architecture is highly expressive, yet simple, and helps avoid adding unnecessary complexity to ACE. The bitmask $\mathbf{b}$, which indicates observed features, is sampled for each training example by first drawing $k \sim \mathcal{U}\{0, d - 1\}$, then choosing $k$ distinct features with uniform probability to be observed. Full experimental details and hyperparameters can be found in the Appendix.

We also consider the scenario in which data features are completely missing, i.e.,~some features are deemed unavailable for particular instances during training and are never part of the observed or unobserved set.\footnote{Features are missing at the per-instance level. For example, this does not mean that the $i^{\text{th}}$ feature is never observed for all training instances.} This allows us to examine the effectiveness of ACE on incomplete datasets, which are common when working with real-world data. When training models with missing data, we simply modify the sets of observed and unobserved indices to remove any indices which have been declared missing. This is a trivial modification and requires no other change to the design or training procedure of ACE. We consider two scenarios where data are missing completely at random at a 10\% and 50\% rate.

\subsubsection{Likelihood Evaluation}

\begin{table*}[t]
\centering
\caption{Test arbitrary conditional log-likelihoods (in nats) for UCI datasets. Higher is better. Likelihood estimates are computed with 20,000 importance samples for \textsc{Power}, \textsc{Gas}, and \textsc{Hepmass}, 10,000 importance samples for \textsc{Miniboone}, and 3,000 importance samples for \textsc{BSDS}. Results for ACFlow and VAEAC are taken from \citet{li2020acflow}. The best model for each dataset and missing rate is shown in bold. Results are averaged over 5 observed masks.}
\label{tab:ac-likelihood-results}

\resizebox{\textwidth}{!}{%
\begin{tabular}{@{}lrrrrrrrrrrrrrrr@{}}
\toprule
             & \multicolumn{3}{c}{\textsc{Power}} & \multicolumn{3}{c}{\textsc{Gas}} & \multicolumn{3}{c}{\textsc{Hepmass}} & \multicolumn{3}{c}{\textsc{Miniboone}} & \multicolumn{3}{c}{BSDS} \\ \cmidrule(l){2-16} 
 Missing Rate &
  \multicolumn{1}{c}{0.0} &
  \multicolumn{1}{c}{0.1} &
  \multicolumn{1}{c}{0.5} &
  \multicolumn{1}{c}{0.0} &
  \multicolumn{1}{c}{0.1} &
  \multicolumn{1}{c}{0.5} &
  \multicolumn{1}{c}{0.0} &
  \multicolumn{1}{c}{0.1} &
  \multicolumn{1}{c}{0.5} &
  \multicolumn{1}{c}{0.0} &
  \multicolumn{1}{c}{0.1} &
  \multicolumn{1}{c}{0.5} &
  \multicolumn{1}{c}{0.0} &
  \multicolumn{1}{c}{0.1} &
  \multicolumn{1}{c}{0.5} \\ \midrule
ACE &
  \textbf{0.631} &
  \textbf{0.633} &
  \textbf{0.600} &
  \textbf{9.643} &
  \textbf{9.526} &
  \textbf{8.530} &
  \textbf{-3.859} &
  \textbf{-4.255} &
  \textbf{-8.133} &
  \textbf{0.310} &
  \textbf{-0.688} &
  \textbf{-5.701} &
  \textbf{86.701} &
  \textbf{86.130} &
  \textbf{80.613} \\
ACE Proposal & 0.583   & 0.573 & 0.542  & 9.484  & 9.348  & 8.183 & -4.417  & -4.796  & -8.497  & -0.241   & -1.328  & -9.169   & 85.228 & 84.204 & 75.767 \\ \midrule
ACFlow       & 0.561   & 0.557  & 0.458  & 8.086  & 7.568  & 5.405 & -8.197  & -7.784  & -10.538 & -0.972   & -5.150  & -9.892   & 81.827 & 80.783 & 75.050 \\
ACFlow+BG    & 0.528   & 0.510  & 0.417  & 7.593  & 7.212  & 4.818 & -6.833  & -9.670  & -10.975 & -1.098   & -3.577  & -10.849  & 81.399 & 79.745 & 73.061 \\
VAEAC        & -0.042  & -0.103 & -0.343 & 2.418  & 2.823  & 1.952 & -10.082 & -10.389 & -11.415 & -3.452   & -4.242  & -9.051   & 74.850 & 74.313 & 66.628 \\ \bottomrule
\end{tabular}%
}

\end{table*}
\begin{table*}[t]
\centering
\caption{Test marginal log-likelihoods (in nats) for UCI datasets. Higher is better. We evaluate the marginal distributions of the first 3, 5, and 10 dimensions of each dataset (\textsc{Power} and \textsc{Gas} don't have 10 features, so the joint likelihood over all features is reported instead). The same number of importance samples are used as in \autoref{tab:ac-likelihood-results}. Results for ACFlow and TAN are taken from \citet{li2020acflow}.
Bold indicates where ACE outperformed ACFlow. Results are averaged over 5 observed masks.}
\label{tab:marginal-likelihood-results}

\resizebox{\textwidth}{!}{%
\begin{tabular}{@{}lrrrrrrrrrrrrrrr@{}}
\toprule
             & \multicolumn{3}{c}{\textsc{Power}} & \multicolumn{3}{c}{\textsc{Gas}} & \multicolumn{3}{c}{\textsc{Hepmass}} & \multicolumn{3}{c}{\textsc{Miniboone}} & \multicolumn{3}{c}{BSDS} \\ \cmidrule(l){2-16} 
 Dimensions &
  \multicolumn{1}{c}{3} &
  \multicolumn{1}{c}{5} &
  \multicolumn{1}{c}{6} &
  \multicolumn{1}{c}{3} &
  \multicolumn{1}{c}{5} &
  \multicolumn{1}{c}{8} &
  \multicolumn{1}{c}{3} &
  \multicolumn{1}{c}{5} &
  \multicolumn{1}{c}{10} &
  \multicolumn{1}{c}{3} &
  \multicolumn{1}{c}{5} &
  \multicolumn{1}{c}{10} &
  \multicolumn{1}{c}{3} &
  \multicolumn{1}{c}{5} &
  \multicolumn{1}{c}{10} \\ \midrule
ACE &
  \textbf{-0.56} &
  \textbf{1.42} &
  \textbf{0.58} &
  \textbf{1.31} &
  \textbf{4.31} &
  \textbf{12.20} &
  \textbf{-4.00} &
  \textbf{-5.91} &
  \textbf{-10.72} &
  \textbf{-2.13} &
  \textbf{-3.80} &
  \textbf{-7.94} &
  \textbf{5.10} &
  \textbf{9.37} &
  \textbf{20.31} \\
ACE Proposal & -0.58   & 1.35   & 0.49   & 1.11   & 3.98  & 11.84  & -4.01   & -5.94   & -10.82  & -2.14    & -3.79   & -7.93    & 5.06   & 9.30   & 20.17  \\ \midrule
ACFlow       & -0.57   & 1.34   & 0.42   & 0.78   & 3.01  & 10.13  & -4.03   & -6.19   & -11.58  & -2.76    & -5.31   & -10.36   & 5.06   & 9.26   & 19.60  \\ \midrule
TAN          & -0.54   & 1.40   & 0.57   & 1.22   & 4.47  & 12.09  & -4.00   & -5.92   & -10.87  & -2.13    & -3.73   & -8.13    & 5.11   & 9.43   & 20.44  \\ \bottomrule
\end{tabular}%
}

\end{table*}

\autoref{tab:ac-likelihood-results} presents the average arbitrary conditional log-likelihoods on held-out test data from models trained with different levels of missing data. During inference, no data is missing and $\mathbf{b}$ is drawn from a Bernoulli distribution with $p=0.5$. Likelihoods are calculated autoregressively as described in \autoref{sec:likelihoods}. The order in which the unobserved one-dimensional conditionals are computed is randomly selected for each instance.

We can draw two key findings from \autoref{tab:ac-likelihood-results}. First, we see that our proposal distribution outperforms ACFlow in all cases. Even this exceptionally simple approach (just a fully-connected network that produces a mixture of Gaussians) can give rise to extremely competitive performance, and we see there are advantages to using decomposed densities and unrestricted network architectures. Second, we find that in every case, the likelihood estimates produced by the energy function are higher than those from the proposal, illustrating the benefits of an energy-based approach which imposes no biases on the shape of the learned distributions.

We also examine the arbitrary marginal distributions learned by ACE, i.e.,~the unconditional distribution over a subset of features. We again test our model against ACFlow, and we additionally compare to Transformation Autoregressive Networks (TAN) \cite{oliva2018transformation}, which are designed only for joint likelihood estimation. A separate TAN model has to be trained for each marginal distribution. While a single ACFlow model can estimate all marginal distributions, \citet{li2020acflow} retrained models specifically for arbitrary marginal estimation. Contrarily, we used the same ACE models when evaluating arbitrary marginals as were used for arbitrary conditionals. Results are provided in \autoref{tab:marginal-likelihood-results}. We find that ACE outperforms ACFlow in all cases and even surpasses TAN most of the time, even though ACFlow and TAN both received special training for marginal likelihood estimation and ACE did not.

\subsubsection{Imputation}
\label{sec:imputation}

\begin{figure*}[t]
    \centering
    \includegraphics[width=\linewidth]{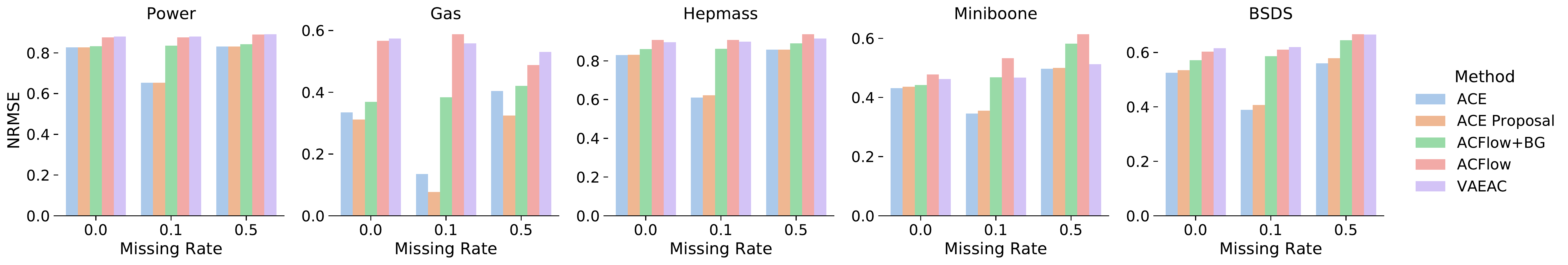}
    \caption{Normalized root-mean-square error (NRMSE) of imputations generated by ACE. Lower is better. NRMSE is computed as the root-mean-square error normalized by the standard deviation of each feature and then averaged across all features. Estimates of energy distribution means are computed with 20,000 importance samples for \textsc{Power} and \textsc{Gas}, 10,000 importance samples for \textsc{Hepmass} and \textsc{Miniboone}, and 3,000 importance samples for \textsc{BSDS}. Results for ACFlow and VAEAC are taken from \citet{li2020acflow}.}
    \label{fig:ac-imputation}
\end{figure*}

We also evaluate ACE for data imputation, where some elements are missing from a dataset completely at random and we seek to infer their values. ACE is naturally applied to this task --- we consider $p(\mathbf{x}_u \mid \mathbf{x}_o)$, where $\mathbf{x}_u$ contains the missing features.

\autoref{fig:ac-imputation} shows the normalized root-mean-square error (NRMSE) on held-out test data. Again, we consider models trained with three different levels of missing data. During inference, $\mathbf{b}$ is drawn from a Bernoulli distribution with $p = 0.5$ for the 0\% and 50\% missing rates and $p = 0.9$ for the 10\% missing rate.\footnote{This means that for the 10\% missing rate, we are imputing fewer values (only 10\% as opposed to 50\%) during inference than with the other two missing rates. This explains the lower errors for the 10\% missing rate that we see in \autoref{fig:ac-imputation}. Even though there is missing data during training, we are estimating the values of fewer values based on a larger amount of observed data, which intuitively should result in more accurate imputations.} The means of the unobserved distributions are used as the imputed values (see \autoref{sec:means}).

As seen in \autoref{fig:ac-imputation}, ACE achieves a lower NRMSE score than ACFlow in all cases (exact numbers are available in the Appendix). These results further validate ACE's ability to accurately model arbitrary conditionals, leading us to again advocate for simple methods with few biases. It is also worth noting that ACE and ACE Proposal do comparably in this imputation task, which estimates the first-order moment of conditional distributions. However, as evidenced in \autoref{tab:ac-likelihood-results}, the energy-based likelihood better captures higher-order moments.

\subsection{High-dimensional Data}

\begin{figure}[t]
    \centering
    \begin{minipage}[t]{0.47\textwidth}
    \centering
    \captionof{table}{Conditional log-likelihoods (LL), joint bits-per-dimension (BPD), and inpainting peak signal-to-noise-ratio (PSNR) for ACE and ACFlow on MNIST.}
    \label{tab:mnist}
    \vskip 0.2cm
    \begin{tabular}{lrrr}
    \toprule
    Method & \multicolumn{1}{c}{LL} & \multicolumn{1}{c}{BPD} & \multicolumn{1}{c}{PSNR} \\ \midrule
    ACE          & \textbf{1043.88} & \textbf{1.42} & \textbf{16.81} \\
    ACE Proposal & 828.60  & 2.11 & 16.70 \\ \midrule
    ACFlow       & 875.87  & 3.09  & 13.75 \\ \bottomrule
    \end{tabular}
\end{minipage}
\hfill
\begin{minipage}[t]{0.51\textwidth}
    \centering
    \captionof{table}{Log-likelihood and imputation results for the UCI Adult dataset. NRMSE measures imputation performance for continuous features and accuracy is used for discrete features.}
    \label{tab:adult}
    \vskip 0.2cm
    \begin{tabular}{lrrr}
    \toprule
    Method & \multicolumn{1}{c}{LL} & \multicolumn{1}{c}{NRMSE} & \multicolumn{1}{c}{Accuracy} \\ \midrule
    ACE          & \textbf{2.38}  & 0.90 & \textbf{0.69}   \\
    ACE Proposal & 2.24  & \textbf{0.89} & \textbf{0.69} \\ \midrule
    VAEAC        & -7.25 & 0.91 & 0.67 \\ \bottomrule
    \end{tabular}%
\end{minipage}
\end{figure}

We examine ACE's ability to scale to general high-dimensional data by considering the flattened MNIST dataset \cite{lecun2010mnist} (i.e.,~each example is a 784-dimensional vector). 
In order to make a fair comparison, we trained a non-convolutional ACFlow model (using the authors' code) on the flattened data as well. The only change to ACE's training procedure is in the distribution of bitmasks that are sampled during training, which we detail in the Appendix.

\begin{figure}[t]
    \centering
    \begin{subfigure}[b]{0.53\linewidth}
    \centering
    \includegraphics[width=\textwidth]{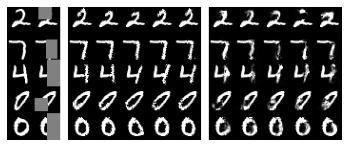}
    \caption{Inpaintings from ACE and ACFlow. Left: Groundtruth and observed pixels. Middle: ACE samples. Right: ACFlow samples.}
    \label{fig:mnist-samples-ac}
    \end{subfigure}
    \hfill
    \begin{subfigure}[b]{0.435\textwidth}
    \centering
    \includegraphics[width=\linewidth]{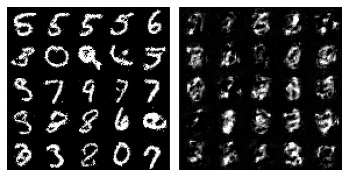}
    \caption{Samples from the joint distribution (i.e.,~all pixels are unobserved). Left: ACE samples. Right: ACFlow samples.}
    \label{fig:mnist-samples-joint}
    \end{subfigure}
    \caption{MNIST samples generated by ACE and ACFlow.}
    \label{fig:mnist-samples}
\end{figure}

Table \ref{tab:mnist} compares ACE and ACFlow in terms of arbitrary conditional log-likelihoods, joint bits-per-dimension (BPD), and peak signal-to-noise-ratio for inpaintings, and we see that ACE outperforms ACFlow for all three metrics. We also note that ACE's BPD of 1.42 is comparable to prior methods for joint density estimation such as TAN, MADE, and MAF, which have reported BPD of 1.19, 1.41, and 1.52 respectively (lower is better), despite the fact that these other methods can \textit{only} model the joint distribution \cite{oliva2018transformation}. Qualitatively, we see that ACE produces more realistic inpaintings (see \autoref{fig:mnist-samples-ac}) and samples from the joint (see \autoref{fig:mnist-samples-joint}) than ACFlow. These findings indicate that ACE's performance scales well to high-dimensional data.

\subsection{Mixed Continuous-Discrete Data}

In order to demonstrate ACE's ability to model data with both continuous and discrete features, we conduct experiments on the UCI Adult dataset\footnote{\url{https://archive.ics.uci.edu/ml/datasets/adult}}, which offers a relatively even balance between the number of continuous and discrete features (see below). We preprocess the data by standardizing continuous features and dropping instances with missing values. We also only keep instances for which the \texttt{native-country} feature is \texttt{United-States}.\footnote{Such instances account for about 91\% of the data, and the other 9\% take on any of 39 possible values. Thus we are able to discard a highly class-imbalanced feature while retaining the vast majority of the data.} The processed data has 6 continuous features and 8 discrete features and is split into train, validation, and test partitions of size 22003, 5501, and 13788 respectively.

We trained an ACE model and VAEAC model (using the authors' publicly available code) on this dataset and evaluated them in terms of likelihoods and imputation. For discrete features, the mode of the learned distribution is imputed (as opposed to the mean). The results are presented in Table \ref{tab:adult}, where we see that ACE outperforms VAEAC on all metrics. We also reiterate that due to its use of the change of variable theorem, ACFlow cannot be trained on this dataset.

\subsection{Importance Sampling Accuracy}

\begin{figure}[t]
    \centering
    \includegraphics[width=\linewidth]{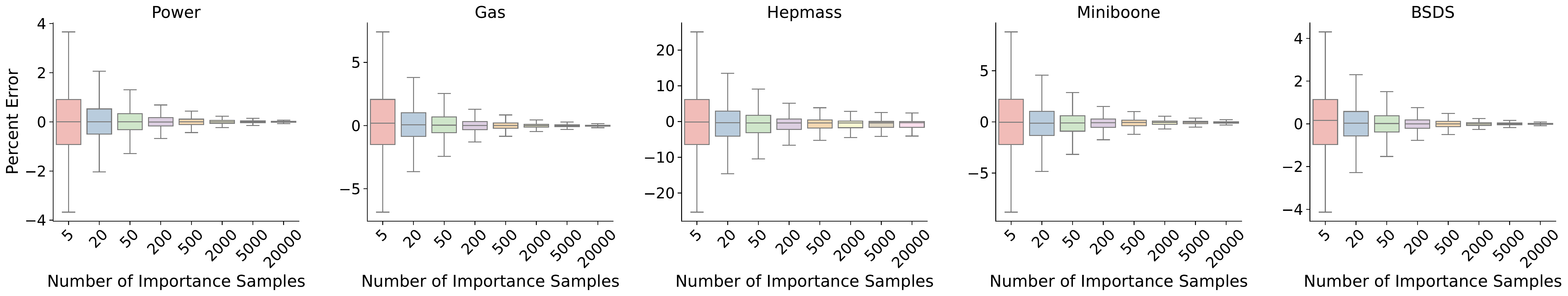}
    \caption{Percent error of normalizing constant estimates obtained with importance sampling when compared to ``true'' values obtained with numerical integration. Estimates grow more accurate as the number of importance samples increases. Whiskers indicate 1.5 times the IQR.}
    \label{fig:is-box}
\end{figure}

In order to better understand the impact of the number of importance samples used when estimating normalizers, we compare the importance sampling estimates from our model to ``true'' estimates obtained via numerical integration (similar to \cite{nash2019autoregressive}). For each UCI dataset, we randomly select 10,000 one-dimensional conditionals (from a validation set that was not seen during training) and integrate the unnormalized energy likelihoods using a trapezoidal rule \cite{pitkin2017lintegrate}. For each conditional, we then also estimate the normalizer with an increasing number of importance samples. \autoref{fig:is-box} shows the percentage errors of the importance sampling estimates. In most cases, a small number of samples (e.g., 20) can already produce relatively accurate estimates, and we see that the estimates grow more accurate as the number of samples increases. 

\section{Conclusion}

In this work, we present a simple approach to modeling all arbitrary conditional distributions $p(\mathbf{x}_u \mid \mathbf{x}_o)$ over a set of covariates. Our method, Arbitrary Conditioning with Energy (ACE), is the first to wholly reduce arbitrary conditioning to one-dimensional conditionals with arbitrary observations and to estimate these with energy functions. By using an energy function to specify densities, ACE can more easily model highly complex distributions, and it can freely use high-capacity networks to model the exponentially many distributions at hand.

Empirically, we find that ACE achieves state-of-the-art performance for arbitrary conditional/marginal density estimation and for data imputation. For a given dataset, all of these results are produced with the same trained model. This high performance does not come at the cost of complexity --- ACE is much simpler than other common approaches, which often require restrictive complexities such as normalizing flow models or networks with specially masked connections \cite{li2020acflow,germain2015made}.

Furthermore, ACE's proposal distribution, a basic mixture of Gaussians, outperforms prior methods, demonstrating that the principle of learning one-dimensional distributions is still powerful when decoupled from energy-based learning. These results emphasize that seemingly complex problems do not necessitate highly complex solutions, and we believe future work on arbitrary density estimation will benefit from similar ideas.

\acksection

This work was supported in part by grants NIH 1R01AA02687901A1 and NSF IIS2133595.

\bibliography{references}

\newpage

\newpage
\appendix

{\LARGE \textbf{Appendix}}

\section{Sampling from ACE}

Sampling the proposal distribution can be performed in an autoregressive fashion where $x_{u_i}$ is sampled from \mbox{$q(x_{u_i} \mid \mathbf{x}_o)$} then added to the observed set, at which point $x_{u_{i + 1}}$ can be sampled. We do this until all unobserved features have been sampled. The pseudocode for this procedure is presented in \autoref{alg:proposal-sample}.

We also want to produce samples that come from the energy function. One drawback of energy-based models is that we are unable to analytically sample the learned distribution. However, there are several methods for obtaining approximate samples. We employ a modification of the proposal sampling procedure such that many proposal samples are drawn at each step, and a single sample is then chosen from that collection based on importance weights. As the number of samples goes to infinity, this is consistent with drawing samples from the energy distribution. The pseudocode for this procedure is presented in \autoref{alg:energy-sample}. We note that this method of sampling is closely related to sampling importance resampling \cite{rubin1988using}.

\section{Algorithms}

For convenience, we provide the procedure for ACE's test-time likelihood evaluation in \autoref{alg:likelihood} and sampling in \autoref{alg:proposal-sample} and \autoref{alg:energy-sample}.

\begin{algorithm}[p]
    \caption{ACE Likelihood Evaluation}
    \label{alg:likelihood}
    \begin{algorithmic}[1]
       \STATE {\bfseries Input:} $\mathbf{x}_o$, $\mathbf{x}_u$, $\mathbf{b}$
       \STATE Set $\mathbf{x}_{cur} = \boldsymbol{\phi}(\mathbf{x}_o;\mathbf{b})$ and $\mathbf{b}_{cur} = \mathbf{b}$
       \STATE Initialize $r = 0$
       \STATE Choose an arbitrary permutation $u^\prime$ of $u$
       \FOR{$u^\prime_i$ {\bfseries in} $u^\prime$}
           \STATE Compute $\log p(x_{u^\prime_i} \mid \mathbf{x}_{cur})$ using Equation 8
           \STATE Set $r = r + \log p(x_{u^\prime_i} \mid \mathbf{x}_{cur})$
           \STATE Set $\mathbf{x}_{cur}[u^\prime_i] = x_{u^\prime_i}$
           \STATE Set $\mathbf{b}_{cur}[u^\prime_i] = 1$
       \ENDFOR
       \STATE {\bfseries Output:} $r$, which contains $\log p(\mathbf{x}_u \mid \mathbf{x}_{o})$
    \end{algorithmic}
\end{algorithm}

\begin{algorithm}[p]
    \caption{ACE Proposal Sampling}
    \label{alg:proposal-sample}
    \begin{algorithmic}[1]
       \STATE {\bfseries Input:} $\mathbf{x}_o$, $\mathbf{b}$, $u$
       \STATE Set $\mathbf{x}_{cur} = \boldsymbol{\phi}(\mathbf{x}_o;\mathbf{b})$ and $\mathbf{b}_{cur} = \mathbf{b}$
       \STATE Choose an arbitrary permutation $u^\prime$ of $u$
       \FOR{$u^\prime_i$ {\bfseries in} $u^\prime$}
           \STATE Sample $x_{u^\prime_i} \sim q(x_{u^\prime_i} \mid \mathbf{x}_{cur}; \mathbf{b}_{cur})$
           \STATE Set $\mathbf{x}_{cur}[u^\prime_i] = x_{u^\prime_i}$
           \STATE Set $\mathbf{b}_{cur}[u^\prime_i] = 1$
       \ENDFOR
       \STATE {\bfseries Output:} $\mathbf{x}_{cur}$, which contains the observed and imputed values
    \end{algorithmic}
\end{algorithm}

\begin{algorithm}[p]
    \caption{ACE Energy Sampling}
    \label{alg:energy-sample}
    \begin{algorithmic}[1]
       \STATE {\bfseries Input:} $\mathbf{x}_o$, $\mathbf{b}$, $u$, $N$
       \STATE Set $\mathbf{x}_{cur} = \boldsymbol{\phi}(\mathbf{x}_o;\mathbf{b})$ and $\mathbf{b}_{cur} = \mathbf{b}$
       \STATE Choose an arbitrary permutation $u^\prime$ of $u$
       \FOR{$u^\prime_i$ {\bfseries in} $u^\prime$}
           \STATE Draw samples $\{x_{u^\prime_i}^{(s)}\}_{s=1}^N$ from $q(x_{u^\prime_i} \mid \mathbf{x}_{cur}; \mathbf{b}_{cur})$
           \STATE Compute importance weights for the $N$ samples, as in Equation 6
           \STATE Draw $x_{u^\prime_i}$ from the $N$ samples according to the importance weights
           \STATE Set $\mathbf{x}_{cur}[u^\prime_i] = x_{u^\prime_i}$
           \STATE Set $\mathbf{b}_{cur}[u^\prime_i] = 1$
       \ENDFOR
       \STATE {\bfseries Output:} $\mathbf{x}_{cur}$, which contains the observed and imputed values
    \end{algorithmic}
\end{algorithm}

\section{Order Consistency}

Because ACE can compute likelihoods by using any permutation of $u$, there are numerous ways to compute $p(\mathbf{x}_u \mid \mathbf{x}_o)$ for a given $\mathbf{x}_u$ and $\mathbf{x}_o$. However, due to inaccuracies in the learned model, we may obtain different results depending on which ordering is used. This phenomenon has surfaced in prior work as well \cite{uria2014deep,germain2015made}, where it has been argued that different orderings can be treated as an advantageous ensemble of models. This perspective is certainly useful, but ideally, our model should give equivalent likelihoods for all orderings. In order to better understand ACE's susceptibility to this problem, as well as how it may be addressed, we do a straightforward experiment in which we fine-tune trained ACE models with an additional loss term that minimizes the variance of $p(\mathbf{x}_u \mid \mathbf{x}_o)$ computed (autoregressively) over 10 permutations of $u$. Intuitively, as the expected variance over all $\mathbf{x}$ and $\mathbf{b}$ goes to zero, the distributions induced by different orderings of $u$ become the same and the model gives the same likelihood regardless of the chosen ordering.

\autoref{tab:consistency} shows the results of these experiments. We find that ACE models can be effectively fine-tuned to produce more consistent likelihoods over different orderings, at almost no cost in performance in terms of the average likelihood. However, we see that if desired, even stronger consistency can be obtained for a slight tradeoff in the average likelihood.

\begin{table}[b]
\centering
\caption{Log-likelihoods after different amounts of consistency fine-tuning. The number after the $\pm$ is the average standard deviation of $p(\mathbf{x}_u \mid \mathbf{x}_o)$ when computed over 1000 randomly chosen orderings. The coefficient refers to the weight of the variance term in the loss during fine-tuning. The 0.0 coefficient refers to the model with which the fine-tuning was initialized.}
\label{tab:consistency}
\vskip 0.2cm
\begin{tabular}{@{}lrrrr@{}}
\toprule
Coefficient & \multicolumn{1}{c}{0.0} & \multicolumn{1}{c}{0.5} & \multicolumn{1}{c}{1.0} & \multicolumn{1}{c}{2.0} \\ \midrule
\textsc{Power}       & $0.622 \pm 0.072$           & $0.622 \pm 0.063$           & $0.620 \pm 0.058$           & $0.619 \pm 0.051$           \\
\textsc{Gas}         & $9.583 \pm 0.513$           & $9.587 \pm 0.457$           & $9.556 \pm 0.420$           & $9.497 \pm 0.376$           \\
\textsc{Hepmass}     & $-3.555 \pm 0.878$    & $-3.669 \pm 0.718$    & $-3.823 \pm 0.621$    & $-4.090 \pm 0.505$    \\ \bottomrule
\end{tabular}
\end{table}

\section{Experimental and Implementation Details}
\label{sec:exp-details}

\begin{table}[p]
\centering
\caption{UCI datasets used in our experiments.}
\label{tab:datasets}
\vskip 0.15in
\begin{tabular}{@{}lrr@{}}
\toprule
Dataset   & \multicolumn{1}{l}{Instances} & Dimensions \\ \midrule
\textsc{Power}     & 1.66M                         & 6          \\
\textsc{Gas}       & 852K                          & 8          \\
\textsc{Hepmass}   & 315K                          & 21         \\
\textsc{Miniboone} & 29.6K                         & 43         \\
BSDS      & 1M                            & 63         \\ \bottomrule
\end{tabular}%
\vskip -0.1in
\end{table}

\begin{figure}[p]
    \centering
    \includegraphics[width=0.5\linewidth]{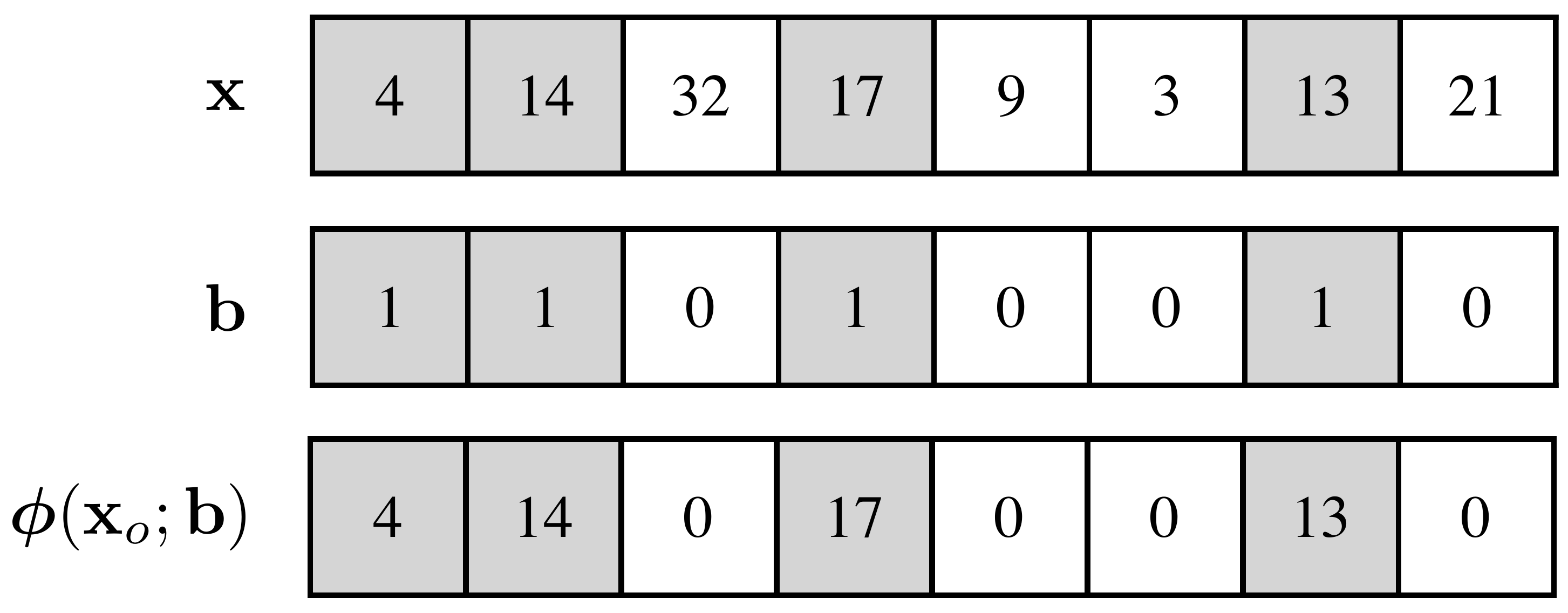}
    \caption{We use a bitmask $\mathbf{b}$ and zero-imputing function $\boldsymbol{\phi}(\cdot ; \mathbf{b})$ to ensure network inputs always have the same shape, regardless of how many features are observed or unobserved. In the figure, shaded cells correspond to observed features.}
    \label{fig:zero-imputation}
\end{figure}

We used a fully-connected residual architecture for both the proposal and energy networks. Each network uses pre-activation residual blocks \cite{he2016identity} and ReLU activations. %

While the energy network only outputs one energy at a time, we can compute energies for every unobserved dimension in parallel by processing them as a batch. A softplus activation is applied to the network's output to ensure energies are nonnegative. We also enforce an upper bound on the energies by manually clipping the network outputs. This is equivalent to setting a lower bound on the unnormalized likelihoods, and we found it improved stability during training. A bound of 30 worked well in our experiments.

During training, we approximate normalizing constants with 20 importance samples from the proposal distribution. Proposal distributions used 10 mixture components, and the minimum allowed scale of each component was 0.001. A small amount of Gaussian noise was added to continuous values in each batch of data during training, as we found it improved stability. The learning rate was linearly annealed over the course of training. We used a warm-up period at the beginning of training where only the proposal network is optimized so that importance sampling does not occur until the proposal is sufficiently similar to the target distribution. \autoref{tab:hyperparameters} gives the hyperparameters that varied between datasets. Evaluations were performed using the weights that produced the highest likelihoods on a set of validation data during training.

All models except for MNIST were trained on two Tesla V100 GPUs, and training time varied from roughly a few hours to a day depending on the dataset. The MNIST model trained on four Tesla V100 GPUs for between one and two days. However, we note that multiple GPUs are not necessary for good results --- we found that state-of-the-art performance can still be achieved by training ACE models on a single GPU with smaller batch sizes.

\subsection{MNIST}

When training on MNIST, images were scaled to the range $[0, 1]$, and the reported likelihoods are evaluated in that space.

For MNIST, we use a different masking scheme during training so that the model learns to inpaint specific types of regions, such as square cutouts. The mask for each example is sampled from a mixture of the following distributions:
\begin{itemize}
    \item \textbf{Bernoulli:} Each pixel is randomly selected to be observed with probability $p=0.5$.
    \item \textbf{Half:} The upper, lower, left, or right half of the image is randomly selected to be observed.
    \item \textbf{Rectangular:} A random rectangle within the image is selected to be unobserved, with the constraint that the area of the rectangle is at least 30\% of the image.
    \item \textbf{Square:} A square with a fourth of the area of the image is randomly selected to be unobserved.
\end{itemize}

During training (but not at test time), each sampled mask was also overlaid with an additional Bernoulli mask for $p \sim \mathcal{U}(0.02, 0.98)$ in order to help simulate the distribution of masks that the model will encounter during the autoregressive procedures it uses during inference. At test time, the extra Bernoulli noise was not used when sampling masks.

ACFlow was trained analogously to ACE, using the authors' code. 

\begin{table}[b]
\centering
\caption{Dataset-specific hyperparameters.}
\label{tab:hyperparameters}
\vskip 0.1cm
\resizebox{\textwidth}{!}{%
\begin{tabular}{@{}lrrrrrrr@{}}
\toprule
\textsc{Hyperparameter} & \multicolumn{1}{l}{\textsc{Power}} & \multicolumn{1}{l}{\textsc{Gas}} & \multicolumn{1}{l}{\textsc{Hepmass}} & \multicolumn{1}{l}{\textsc{Miniboone}} & \multicolumn{1}{l}{BSDS} & \multicolumn{1}{l}{\textsc{Adult}} & \multicolumn{1}{l}{MNIST} \\ \midrule
Dropout                    & 0.2     & 0.0    & 0.2    & 0.5    & 0.2 & 0.5 & 0.2  \\
MSE Penalty Coef.          & 1.0     & 0.0    & 0.0    & 0.0    & 0.0 & 1.0 & 0.0  \\
Training Steps             & 1600000 & 1000000 & 1000000 & 15000 & 1000000 & 40000 & 800000 \\
Warm-up Steps              & 5000    & 5000   & 5000   & 100   & 5000  & 2500 & 100000 \\
Training Noise Scale       & 0.003   & 0.001  & 0.001  & 0.005  & 0.001 & 0.005 & 0.01 \\
Learning Rate              & 0.0001  & 0.001  & 0.0005  & 0.001  & 0.001 & 0.0005 & 0.0002 \\
Batch Size                 & 512     & 2048  & 2048  & 2048  & 2048 & 1024 & 64 \\
Proposal Hidden Dim. & 512     & 512    & 512    & 512    & 1024 & 512 & 1024  \\
Proposal Res. Blocks       & 4     & 4  & 4  & 4  & 4 & 4 & 5 \\
Proposal Latent Output Dim. & 64  & 64  & 64  & 64  & 64 & 64 & 128 \\
\bottomrule
\end{tabular}%
}
\end{table}

\section{Results}

Table \ref{tab:ac-ll-appendix} presents the full UCI likelihood results with standard deviations. In the main text, the imputation results are presented as a graph. We give the values that generated the graph, along with standard deviations, in Table \ref{tab:imputation-appendix}.

\begin{figure}[p]
    \centering
    \begin{minipage}[t]{0.4485\textwidth}
    \centering
\captionof{table}{Arbitrary conditional log-likelihood results for UCI datasets. Standard deviation is over 5 trials with different observed masks.}
\label{tab:ac-ll-appendix}
\vskip 0.2cm
\resizebox{\textwidth}{!}{%
\begin{tabular}{@{}llrrr@{}}
\toprule
Dataset   & Method       & \multicolumn{1}{l}{Missing Rate} & \multicolumn{1}{l}{LL Mean} & \multicolumn{1}{l}{LL Std.} \\ \midrule
Power     & ACE          & 0.0                              & 0.631                       & 0.002                       \\
Power     & ACE Proposal & 0.0                              & 0.583                       & 0.003                       \\
Power     & ACFlow       & 0.0                              & 0.561                       & 0.003                       \\
Power     & ACFlow+BG    & 0.0                              & 0.528                       & 0.003                       \\
Power     & VAEAC        & 0.0                              & -0.042                      & 0.002                       \\
Power     & ACE          & 0.1                              & 0.633                       & 0.003                       \\
Power     & ACE Proposal & 0.1                              & 0.573                       & 0.003                       \\
Power     & ACFlow       & 0.1                              & 0.557                       & 0.003                       \\
Power     & ACFlow+BG    & 0.1                              & 0.510                       & 0.003                       \\
Power     & VAEAC        & 0.1                              & -0.103                      & 0.005                       \\
Power     & ACE          & 0.5                              & 0.600                       & 0.003                       \\
Power     & ACE Proposal & 0.5                              & 0.542                       & 0.003                       \\
Power     & ACFlow       & 0.5                              & 0.458                       & 0.005                       \\
Power     & ACFlow+BG    & 0.5                              & 0.417                       & 0.005                       \\
Power     & VAEAC        & 0.5                              & -0.343                      & 0.004                       \\
Gas       & ACE          & 0.0                              & 9.643                       & 0.005                       \\
Gas       & ACE Proposal & 0.0                              & 9.484                       & 0.005                       \\
Gas       & ACFlow       & 0.0                              & 8.086                       & 0.010                       \\
Gas       & ACFlow+BG    & 0.0                              & 7.593                       & 0.011                       \\
Gas       & VAEAC        & 0.0                              & 2.418                       & 0.006                       \\
Gas       & ACE          & 0.1                              & 9.526                       & 0.007                       \\
Gas       & ACE Proposal & 0.1                              & 9.348                       & 0.007                       \\
Gas       & ACFlow       & 0.1                              & 7.568                       & 0.005                       \\
Gas       & ACFlow+BG    & 0.1                              & 7.212                       & 0.008                       \\
Gas       & VAEAC        & 0.1                              & 2.823                       & 0.009                       \\
Gas       & ACE          & 0.5                              & 8.530                       & 0.007                       \\
Gas       & ACE Proposal & 0.5                              & 8.183                       & 0.005                       \\
Gas       & ACFlow       & 0.5                              & 5.405                       & 0.008                       \\
Gas       & ACFlow+BG    & 0.5                              & 4.818                       & 0.009                       \\
Gas       & VAEAC        & 0.5                              & 1.952                       & 0.023                       \\
Hepmass   & ACE          & 0.0                              & -3.859                      & 0.005                       \\
Hepmass   & ACE Proposal & 0.0                              & -4.417                      & 0.005                       \\
Hepmass   & ACFlow       & 0.0                              & -8.197                      & 0.008                       \\
Hepmass   & ACFlow+BG    & 0.0                              & -6.833                      & 0.006                       \\
Hepmass   & VAEAC        & 0.0                              & -10.082                     & 0.010                       \\
Hepmass   & ACE          & 0.1                              & -4.255                      & 0.003                       \\
Hepmass   & ACE Proposal & 0.1                              & -4.796                      & 0.003                       \\
Hepmass   & ACFlow       & 0.1                              & -7.784                      & 0.006                       \\
Hepmass   & ACFlow+BG    & 0.1                              & -9.670                      & 0.007                       \\
Hepmass   & VAEAC        & 0.1                              & -10.389                     & 0.005                       \\
Hepmass   & ACE          & 0.5                              & -8.133                      & 0.007                       \\
Hepmass   & ACE Proposal & 0.5                              & -8.497                      & 0.006                       \\
Hepmass   & ACFlow       & 0.5                              & -10.538                     & 0.006                       \\
Hepmass   & ACFlow+BG    & 0.5                              & -10.975                     & 0.006                       \\
Hepmass   & VAEAC        & 0.5                              & -11.415                     & 0.012                       \\
Miniboone & ACE          & 0.0                              & 0.310                       & 0.054                       \\
Miniboone & ACE Proposal & 0.0                              & -0.241                      & 0.056                       \\
Miniboone & ACFlow       & 0.0                              & -0.972                      & 0.022                       \\
Miniboone & ACFlow+BG    & 0.0                              & -1.098                      & 0.032                       \\
Miniboone & VAEAC        & 0.0                              & -3.452                      & 0.067                       \\
Miniboone & ACE          & 0.1                              & -0.688                      & 0.046                       \\
Miniboone & ACE Proposal & 0.1                              & -1.328                      & 0.057                       \\
Miniboone & ACFlow       & 0.1                              & -5.150                      & 0.053                       \\
Miniboone & ACFlow+BG    & 0.1                              & -3.577                      & 0.057                       \\
Miniboone & VAEAC        & 0.1                              & -4.242                      & 0.071                       \\
Miniboone & ACE          & 0.5                              & -5.701                      & 0.050                       \\
Miniboone & ACE Proposal & 0.5                              & -9.169                      & 0.083                       \\
Miniboone & ACFlow       & 0.5                              & -9.892                      & 0.084                       \\
Miniboone & ACFlow+BG    & 0.5                              & -10.849                     & 0.105                       \\
Miniboone & VAEAC        & 0.5                              & -9.051                      & 0.079                       \\
BSDS      & ACE          & 0.0                              & 86.701                      & 0.008                       \\
BSDS      & ACE Proposal & 0.0                              & 85.228                      & 0.009                       \\
BSDS      & ACFlow       & 0.0                              & 81.827                      & 0.007                       \\
BSDS      & ACFlow+BG    & 0.0                              & 81.399                      & 0.008                       \\
BSDS      & VAEAC        & 0.0                              & 74.850                      & 0.005                       \\
BSDS      & ACE          & 0.1                              & 86.130                      & 0.022                       \\
BSDS      & ACE Proposal & 0.1                              & 84.204                      & 0.020                       \\
BSDS      & ACFlow       & 0.1                              & 80.783                      & 0.018                       \\
BSDS      & ACFlow+BG    & 0.1                              & 79.745                      & 0.017                       \\
BSDS      & VAEAC        & 0.1                              & 74.313                      & 0.015                       \\
BSDS      & ACE          & 0.5                              & 80.613                      & 0.027                       \\
BSDS      & ACE Proposal & 0.5                              & 75.767                      & 0.131                       \\
BSDS      & ACFlow       & 0.5                              & 75.050                      & 0.010                       \\
BSDS      & ACFlow+BG    & 0.5                              & 73.061                      & 0.015                       \\
BSDS      & VAEAC        & 0.5                              & 66.628                      & 0.029                       \\ \bottomrule
\end{tabular}
}

\end{minipage}
\hfill
\begin{minipage}[t]{0.52\textwidth}
    \centering
\captionof{table}{Imputation results for UCI datasets. Standard deviation is over 5 trials with different observed masks.}
\label{tab:imputation-appendix}
\vskip 0.2cm
\resizebox{\textwidth}{!}{%
\begin{tabular}{@{}llrrr@{}}
\toprule
Dataset   & Method       & \multicolumn{1}{l}{Missing Rate} & \multicolumn{1}{l}{NRMSE Mean} & \multicolumn{1}{l}{NRMSE Std.} \\ \midrule
Power     & ACE          & 0.0                              & 0.828                       & 0.002                       \\
Power     & ACE Proposal & 0.0                              & 0.828                       & 0.002                       \\
Power     & ACFlow       & 0.0                              & 0.877                       & 0.001                       \\
Power     & ACFlow+BG    & 0.0                              & 0.833                       & 0.002                       \\
Power     & VAEAC        & 0.0                              & 0.880                       & 0.001                       \\
Power     & ACE          & 0.1                              & 0.653                       & 0.000                       \\
Power     & ACE Proposal & 0.1                              & 0.653                       & 0.000                       \\
Power     & ACFlow       & 0.1                              & 0.877                       & 0.002                       \\
Power     & ACFlow+BG    & 0.1                              & 0.836                       & 0.002                       \\
Power     & VAEAC        & 0.1                              & 0.881                       & 0.003                       \\
Power     & ACE          & 0.5                              & 0.831                       & 0.000                       \\
Power     & ACE Proposal & 0.5                              & 0.831                       & 0.000                       \\
Power     & ACFlow       & 0.5                              & 0.890                       & 0.000                       \\
Power     & ACFlow+BG    & 0.5                              & 0.843                       & 0.001                       \\
Power     & VAEAC        & 0.5                              & 0.892                       & 0.002                       \\
Gas       & ACE          & 0.0                              & 0.335                       & 0.027                       \\
Gas       & ACE Proposal & 0.0                              & 0.312                       & 0.033                       \\
Gas       & ACFlow       & 0.0                              & 0.567                       & 0.050                       \\
Gas       & ACFlow+BG    & 0.0                              & 0.369                       & 0.016                       \\
Gas       & VAEAC        & 0.0                              & 0.574                       & 0.033                       \\
Gas       & ACE          & 0.1                              & 0.135                       & 0.014                       \\
Gas       & ACE Proposal & 0.1                              & 0.077                       & 0.000                       \\
Gas       & ACFlow       & 0.1                              & 0.588                       & 0.025                       \\
Gas       & ACFlow+BG    & 0.1                              & 0.384                       & 0.018                       \\
Gas       & VAEAC        & 0.1                              & 0.558                       & 0.047                       \\
Gas       & ACE          & 0.5                              & 0.404                       & 0.052                       \\
Gas       & ACE Proposal & 0.5                              & 0.325                       & 0.000                       \\
Gas       & ACFlow       & 0.5                              & 0.488                       & 0.030                       \\
Gas       & ACFlow+BG    & 0.5                              & 0.421                       & 0.016                       \\
Gas       & VAEAC        & 0.5                              & 0.531                       & 0.036                       \\
Hepmass   & ACE          & 0.0                              & 0.830                       & 0.001                       \\
Hepmass   & ACE Proposal & 0.0                              & 0.832                       & 0.001                       \\
Hepmass   & ACFlow       & 0.0                              & 0.909                       & 0.000                       \\
Hepmass   & ACFlow+BG    & 0.0                              & 0.861                       & 0.001                       \\
Hepmass   & VAEAC        & 0.0                              & 0.896                       & 0.001                       \\
Hepmass   & ACE          & 0.1                              & 0.610                       & 0.000                       \\
Hepmass   & ACE Proposal & 0.1                              & 0.623                       & 0.000                       \\
Hepmass   & ACFlow       & 0.1                              & 0.908                       & 0.001                       \\
Hepmass   & ACFlow+BG    & 0.1                              & 0.863                       & 0.001                       \\
Hepmass   & VAEAC        & 0.1                              & 0.899                       & 0.000                       \\
Hepmass   & ACE          & 0.5                              & 0.858                       & 0.000                       \\
Hepmass   & ACE Proposal & 0.5                              & 0.858                       & 0.000                       \\
Hepmass   & ACFlow       & 0.5                              & 0.938                       & 0.000                       \\
Hepmass   & ACFlow+BG    & 0.5                              & 0.890                       & 0.000                       \\
Hepmass   & VAEAC        & 0.5                              & 0.915                       & 0.001                       \\
Miniboone & ACE          & 0.0                              & 0.432                       & 0.003                       \\
Miniboone & ACE Proposal & 0.0                              & 0.436                       & 0.004                       \\
Miniboone & ACFlow       & 0.0                              & 0.478                       & 0.004                       \\
Miniboone & ACFlow+BG    & 0.0                              & 0.442                       & 0.001                       \\
Miniboone & VAEAC        & 0.0                              & 0.462                       & 0.002                       \\
Miniboone & ACE          & 0.1                              & 0.346                       & 0.001                       \\
Miniboone & ACE Proposal & 0.1                              & 0.355                       & 0.000                       \\
Miniboone & ACFlow       & 0.1                              & 0.533                       & 0.005                       \\
Miniboone & ACFlow+BG    & 0.1                              & 0.468                       & 0.003                       \\
Miniboone & VAEAC        & 0.1                              & 0.467                       & 0.004                       \\
Miniboone & ACE          & 0.5                              & 0.497                       & 0.000                       \\
Miniboone & ACE Proposal & 0.5                              & 0.500                       & 0.000                       \\
Miniboone & ACFlow       & 0.5                              & 0.614                       & 0.004                       \\
Miniboone & ACFlow+BG    & 0.5                              & 0.582                       & 0.007                       \\
Miniboone & VAEAC        & 0.5                              & 0.513                       & 0.004                       \\
BSDS      & ACE          & 0.0                              & 0.525                       & 0.000                       \\
BSDS      & ACE Proposal & 0.0                              & 0.535                       & 0.000                       \\
BSDS      & ACFlow       & 0.0                              & 0.603                       & 0.000                       \\
BSDS      & ACFlow+BG    & 0.0                              & 0.572                       & 0.000                       \\
BSDS      & VAEAC        & 0.0                              & 0.615                       & 0.000                       \\
BSDS      & ACE          & 0.1                              & 0.389                       & 0.000                       \\
BSDS      & ACE Proposal & 0.1                              & 0.407                       & 0.000                       \\
BSDS      & ACFlow       & 0.1                              & 0.610                       & 0.000                       \\
BSDS      & ACFlow+BG    & 0.1                              & 0.586                       & 0.001                       \\
BSDS      & VAEAC        & 0.1                              & 0.620                       & 0.000                       \\
BSDS      & ACE          & 0.5                              & 0.560                       & 0.000                       \\
BSDS      & ACE Proposal & 0.5                              & 0.579                       & 0.000                       \\
BSDS      & ACFlow       & 0.5                              & 0.667                       & 0.001                       \\
BSDS      & ACFlow+BG    & 0.5                              & 0.645                       & 0.000                       \\
BSDS      & VAEAC        & 0.5                              & 0.666                       & 0.001                       \\ \bottomrule
\end{tabular}
}

\end{minipage}
\end{figure}

\end{document}